\documentclass[a4paper,fleqn]{cas-sc}
\usepackage[authoryear]{natbib}
\usepackage{graphicx}

\begin{document}
\let\WriteBookmarks\relax
\def\floatpagepagefraction{1}
\def\textpagefraction{.001}

\shorttitle{Cross-polarization pairing for label-free few-shot thin-section classification}
\shortauthors{B. Yan}

\title[mode=title]{Cross-polarization pairing for label-free few-shot rock thin-section classification}

\author[1]{Bingchen Yan}
\cormark[1]
\ead{yanbingchen132@126.com}
\affiliation[1]{organization={School of Computer Science and Information Security, Guilin University of Electronic Technology},
            city={Guilin}, postcode={541004}, country={China}}

\cortext[1]{Corresponding author}

\begin{abstract}
Few-shot classifiers of rock thin sections are meta-trained on many rock types that a petrographer has identified slide by slide, and this identification is the costly step that new collections lack. We study few-shot recognition of unseen rock types when no training slide is identified, and the only information available is what the microscope records anyway: which slide an image comes from, and whether it was taken in plane-polarized (PPL) or cross-polarized (XPL) light. We first show why generic unsupervised pseudo-tasks fit this setting poorly: the positives they find without labels are either wrong or near-duplicates of the anchor, so they teach the model little. Polarized-light microscopy provides positives that avoid both problems. The PPL and XPL views of one field show the same grains, so they are correct by construction, yet they look very different. We propose cross-polarization pairing (XPP), which treats each slide as a pseudo-class and pairs its PPL and XPL views of the same field location, so that matching must rely on grain shape and texture rather than on colour. On the Nanjing University collection of 108 rock types, with support and query images taken from different slides, XPP is the best label-free method on average over three random splits, and matches a supervised model trained on 78 identified rock types. Controls show that colour augmentation and other acquisition positives do not reproduce the gain. We also show that the image-level protocol of earlier work is saturated, and propose a cross-slide protocol in its place.
\end{abstract}


\maketitle

\section{Introduction}\label{sec:1}
Thin-section identification is the routine first step of petrographic analysis, and automating it has attracted steady work in deep learning. Because rare rock types are photographed only a few times, recent studies cast the task as few-shot learning: a model meta-trained on many rock types must recognize a new type from one to five images \citep{wang2025rock}. These models reach 96--99\% 5-way 5-shot accuracy on the Nanjing University teaching collection. These figures come from an image-level protocol, in which support and query images can show the same field of view. Under that protocol our label-free method also reaches 98\%, and even untrained ImageNet features reach 97\% (Section~\ref{sec:6.2}). We therefore evaluate with support and query images from different slides, which lowers all accuracies to about 85--90\% and measures recognition of a rock type rather than of a field of view.

That accuracy rests on an assumption the field rarely states. Meta-training needs many rock types whose slides a petrographer has already identified, and identification is the expensive step: it takes an expert, it is subjective for transitional rocks, and it is exactly what a new core, a new outcrop campaign or an old uncatalogued slide archive lacks. What such collections do have is acquisition metadata. Every digital capture records which slide is on the stage, whether the analyzer is in (cross-polarized, XPL) or out (plane-polarized, PPL), and, for rotation series, the rotation angle.

We therefore study a \textbf{weakly annotated} setting: the training images carry acquisition metadata but no rock identification, and the few-shot test is on rock types never seen in training. The setting is the unsupervised few-shot learning (UFSL) problem of the machine-learning literature, with one difference that turns out to matter: the metadata are free, reliable, and specific to how thin sections are imaged.

Generic UFSL methods transfer poorly to this setting, and the reason determines what a better method needs. UFSL methods build pseudo few-shot tasks from unlabeled images, either from augmentations of single images or by grouping images through clustering. Oracle experiments show that the damaging error is the false positive, a sample placed in a pseudo-class it does not belong to, while false negatives (two pseudo-classes of the same true class) cost almost nothing. The false positives cannot be filtered without labels, because the useful positives are precisely those the current features do not recognize as similar.

Thin sections offer a way out that natural images do not. The PPL and XPL photographs of one field show the same grains under two illuminations: interference colours and extinction in one, relief, absorption colour and pleochroism in the other. The pair is the same rock by construction, and its two views look very different (cosine 0.69 with ImageNet features, below the 0.77 of two slides of the same rock type). Pairing them forces a model to match what survives the change of light, which is grain shape, size distribution and texture.

The contribution is thus twofold. Computationally, we diagnose why pseudo-task construction fails without labels and replace label-free positive selection with positives defined by the imaging process. Geoscientifically, we show that the physics of polarized-light microscopy, the change from interference colours to relief and absorption colour over the same grains, is enough to learn rock-type features from slide archives that have never been identified.

Our contributions are:

\begin{enumerate}
  \item \textbf{Which positives are useful.} Oracle experiments show that false positives, not false negatives, limit generic pseudo-tasks, and that label-free selection cannot remove them. On thin sections, a measurement of candidate positive sources shows that only the PPL view of the same field is both correct and less similar to the anchor than another slide of the same rock type.
  \item \textbf{Cross-polarization pairing (XPP).} Pseudo-tasks whose shots pair PPL and XPL views of the same field location. No rock identification is used at any point of training. Under a cross-slide protocol on the primary split, XPP gains 4.4 points over ImageNet features and 2.0 over instance training, and matches a supervised model trained on 78 identified rock types. Averaged over three random splits, the gains are 4.1 and 2.4 points, and XPP remains on par with supervision.
  \item \textbf{Controls and an evaluation protocol.} Strong colour augmentation, other-region and other-angle positives all fail to reproduce the gain, isolating the effect of the polarization pair. We evaluate with support and query from different slides, show that the image-level protocol is saturated, and report variation across random splits.
\end{enumerate}

\section{Related work}\label{sec:2}
\textbf{Deep learning for thin sections.} CNN classifiers of thin-section images range from small sets of sedimentary or igneous rocks to more than a hundred types, often fusing PPL and XPL views \citep{su2020rock,polat2021automatic,polat2021classification,zhu2024rocks2net}. A critical review of rock-type classification points to recurring pitfalls in experimental design, in particular test sets that contain patches of images or samples also used in training \citep{mendes2024rocktype}, and models trained on small collections can rely on cracks, cement or scale bars rather than on framework grains \citep{zheng2024explainable}. Few-shot formulations meta-train on identified rock types; \citet{wang2025rock} add cross-attention PPL--XPL fusion and feature selection to a prototypical network and report 99.16\% 5-way 5-shot accuracy on the Nanjing collection. At the grain scale, the LITHOS benchmark \citep{chacon2025lithos} provides 211\,604 paired PPL and XPL patches with 105\,802 expert-annotated grains of 25 minerals, and shows that a dual-encoder transformer fusing the two polarizations outperforms single-polarization models. All of these methods learn from identified samples and use the PPL--XPL pair as a joint input. We instead use the pair as a source of label-free positives, and recognize rock types unseen in training.

\textbf{Label scarcity in rock imaging.} Other work reduces the cost of annotation in digital rock analysis. Self-supervised pre-training (SimCLR, MoCo-v2, BYOL, NNCLR) on unlabeled micro-CT images improves rock image classification over training from scratch \citep{nguyen2024ssl}, and a self-supervised Vision Transformer segments FIB-SEM images of shale without pixel annotation \citep{mohammad2025ssl}. Diffusion models synthesize rock images with matching masks to augment small segmentation sets \citep{sun2026rocksdm}, and an edge-aware path improves segmentation when training data are limited \citep{wang2025edge}. These studies rely on generic augmentations or synthetic data and target segmentation or the classes seen in pre-training. To our knowledge, none addresses few-shot recognition of unseen rock types, or derives positives from the polarized-light imaging process.

\textbf{Unsupervised few-shot learning.} UFSL methods construct pseudo-tasks from unlabeled data. One family treats every image as its own class and uses augmentations as shots (UMTRA, \citealp{khodadadeh2019unsupervised}; ProtoCLR, \citealp{medina2020self}). Another groups images into pseudo-classes by clustering (CACTUs, \citealp{hsu2018unsupervised}; PsCo, \citealp{jang2023unsupervised}; PETD, \citealp{zhao2025unsupervised}). PETD progressively replaces augmented views with co-clustered neighbours to increase task diversity. Section~\ref{sec:4} studies exactly this replacement and shows that its benefit is limited by false positives.

\textbf{Positives from metadata and feature-space positives.} In medical imaging, multiple images of the same patient serve as positives for contrastive pre-training (MICLe, \citealp{azizi2024big}). Hard positives can also be synthesized by extrapolating features \citep{zhu2021improving}. We differ in two ways: our positives come from the physics of polarized-light microscopy (rotation series and analyzer state), and we motivate them by a measured failure of label-free selection rather than by assumption. Feature extrapolation of a single image cannot supply the cross-instance variation that our diagnosis identifies as missing (Appendix~\ref{app:diag}).

\section{Problem setting and dataset}\label{sec:3}
\subsection{Weakly annotated few-shot setting}\label{sec:3.1}
The training set is a collection of slides from rock types $C_\mathrm{train}$. For each image we know only its slide identifier $s$, its illumination mode $m \in \{\mathrm{PPL}, \mathrm{XPL}\}$ and, for XPL, its rotation angle $\theta$. We never use the rock type of a training slide, for model fitting or for model selection.

Model selection uses a validation set of rock types $C_\mathrm{val}$ that are disjoint from training. Evaluation is $N$-way $K$-shot classification on rock types $C_\mathrm{test}$, disjoint from both. All three metadata fields are produced by the acquisition itself: a slide is on the stage, the analyzer is in or out, the stage or the polarizers are turned. None requires petrographic judgement, so the setting applies to any slide archive that was digitized with its capture log.

\subsection{Dataset}\label{sec:3.2}
We use the Nanjing University teaching thin-section photographs \citep{lai2020photomicrograph}, the same collection as \citet{wang2025rock}. Counts below are from the released archives.

\begin{table}[t]
\centering\small
\caption{Nanjing University thin-section collection (counts from released archives).}
\label{tab:dataset}
\begin{tabular}{lccc}
\toprule
Rock class & Rock types & Slides & Images \\ \midrule
Igneous & 40 & 120 & 963 \\
Metamorphic & 40 & 120 & 972 \\
Sedimentary & 28 & 84 & 699 \\
Total & 108 & 324 & 2634 \\
\bottomrule
\end{tabular}
\end{table}

Each rock type has three slides. Each slide has eight photographs of one field of view, taken within about five minutes. Some slides have a ninth photograph and one has seven. Image 1 is PPL and images 2--8 are XPL of the same field (checked visually on one igneous, one metamorphic and one sedimentary slide). The XPL frames are geometrically aligned with each other, and only interference colours and extinction change; \citet{wang2025rock} describe them as a rotation series at 15$^\circ$ steps. Our method uses only the fact that they show the same field, not the angle values. The optional ninth image shows a magenta background, consistent with XPL with an inserted compensator plate; its acquisition mode is not documented, so we exclude it.

\textbf{Preprocessing.} Every image carries a burnt-in scale bar in the lower-right corner, and its value differs between slides (for example 500~$\mu$m and 200~$\mu$m), so magnification varies across slides. A model could use the bar as a slide-identity shortcut, as observed for supervised classifiers by \citet{zheng2024explainable}, so we mask the corner before cropping.

\begin{figure*}[t]
\centering
\includegraphics[width=1.00\textwidth]{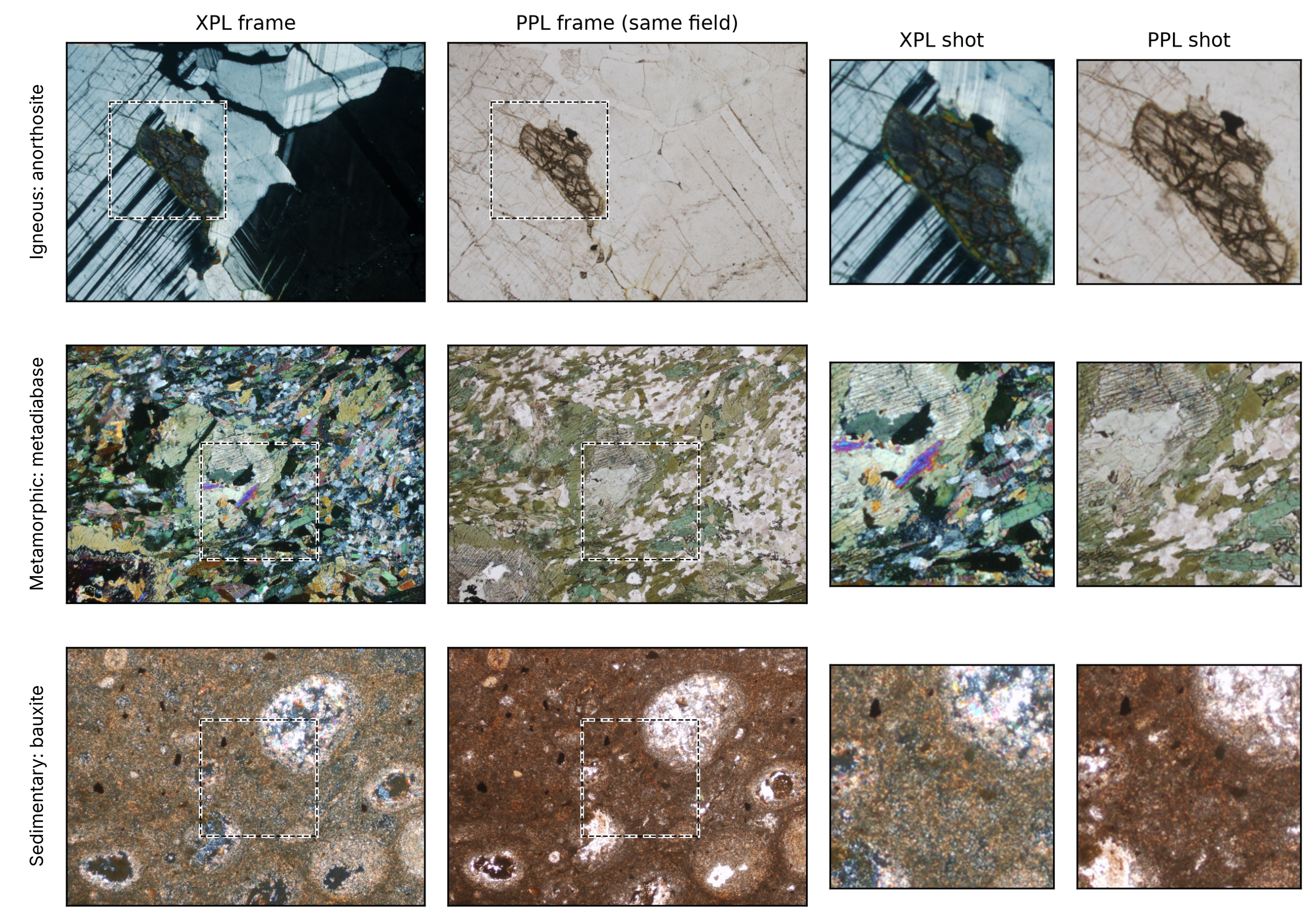}
\caption{Aligned XPL and PPL frames of one field for an igneous (anorthosite), a metamorphic (metadiabase) and a sedimentary (bauxite) slide, after preprocessing. Dashed box: one XPP anchor box; right columns: two 224-px shots it yields. Interference colours and extinction (XPL) are replaced by relief and absorption colour (PPL), while every grain boundary stays in place.}
\label{fig:pairs}
\end{figure*}

File names encode the rock name. We replace every name with an opaque slide hash before training, so the training code sees only (slide, mode, angle).

\subsection{Splits and evaluation protocols}\label{sec:3.3}
After preprocessing, 322 slides remain (one slide lacks its PPL image and one has XPL frames of unequal size). Following \citet{wang2025rock}, rock types are split 78 / 10 / 20 into train, validation and test. The \textbf{primary split} is a random permutation of rock types with seed 0, fixed before any experiment; all main results use it. Two further random splits (seeds 1 and 2) are reported in Section~\ref{sec:6.5}. We report two protocols:

\begin{itemize}
  \item \textbf{Image-level} (as in prior work): support and query images of a class may come from the same slide, and therefore from the same field of view.
  \item \textbf{Cross-slide} (ours, primary): in each test episode, a class's support images come from one slide and its queries are all XPL images of its other slides (up to 14). This measures recognition of a rock type rather than re-identification of a field of view. It follows the requirement that test data must not share images or samples with the data used to build the model \citep{mendes2024rocktype}; in the few-shot setting, the support plays the role of training data.
\end{itemize}

Episodes are 5-way 1-shot and 5-shot, 2000 fixed test episodes; the 95\% confidence interval over episodes is about $\pm$0.5 points. Every trained configuration is run with three seeds, and we report mean $\pm$ standard deviation over seeds.

\section{Which positives are useful?}\label{sec:4}
A few-shot pseudo-task needs positives: samples that the model must treat as the same class as an anchor. Two errors are possible. A false positive places a sample of another class in the pseudo-class; a false negative splits one true class into two pseudo-classes. This section asks which of the two matters, whether it can be avoided without labels, and which positives the imaging of thin sections provides for free.

\subsection{Lessons from generic pseudo-tasks}\label{sec:4.1}
We first ran oracle experiments on PETD, a recent clustering-based method, on miniImageNet, which provides the class labels that such experiments need (details in Appendix~\ref{app:diag}). Three findings shape our method.

\textit{False positives dominate.} About 90\% of the neighbours that PETD injects into a pseudo-class belong to another class. Removing them with an oracle adds 3.1 points of 1-shot and 4.5 points of 5-shot accuracy, whereas removing false negatives changes accuracy by less than 0.4 points.

\textit{They cannot be filtered without labels.} Six label-free selectors tried in training raise the precision of injected neighbours to at most 0.39 and move 1-shot accuracy by at most 0.3 points; graph diffusion, tested offline, does not exceed 0.38. The correct neighbours these selectors find are near-duplicates of the anchor (cosine 0.72, against 0.55 for a random true same-class sample). A selector that relies on the model's own similarity returns what the model already finds similar, so the informative positives are exactly those it misses.

\textit{Requirement.} A useful positive must therefore be correct and dissimilar at the same time, and must come from a source other than the model's own features. False negatives, in contrast, can be accepted at little cost.

\subsection{Candidate positive sources on thin sections}\label{sec:4.2}
Thin-section acquisition provides several positives that are correct by construction: other rotation angles of the same field, other regions of the same image, and the PPL view of the same field. We compare them with two label-free alternatives (augmentation and feature nearest neighbours) and with the oracle positive that a supervised model learns from: another slide of the same rock type. For each source we measure, with ImageNet-pretrained ResNet-50 features, its precision (same rock type) and its mean cosine to the anchor. The measurement uses all 323 slides with complete PPL and XPL frames (108 rock types), with 3 anchor crops per slide (crop side 0.4 of the short side, images resized to 1024~px, scale-bar corner excluded), giving 969 pairs per row.

\begin{table}[t]
\centering\small
\caption{Candidate positive sources on thin sections, measured with ImageNet ResNet-50 features (323 slides, 108 rock types, 969 pairs per row).}
\label{tab:audit}
\begin{tabular}{lccc}
\toprule
Positive source (thin sections) & Needs labels? & Precision & Mean cosine \\ \midrule
Augmentation of the same crop & no & 1 & 0.931 \\
Feature nearest neighbour, other slides & no & 0.348 & 0.860 \\
XPL at another angle, same location & no & 1 & 0.857 \\
Non-overlapping region, same image & no & 1 & 0.840 \\
XPL at another angle, another region & no & 1 & 0.826 \\
PPL $\leftrightarrow$ XPL, same location & no & 1 & 0.689 \\
PPL $\leftrightarrow$ XPL, another region & no & 1 & 0.656 \\
Same rock type, another slide (oracle) & yes & 1 & 0.768 \\
Random other rock type (reference) & -- & 0 & 0.636 \\
\bottomrule
\end{tabular}
\end{table}
\begin{figure}[t]
\centering
\includegraphics[width=0.80\linewidth]{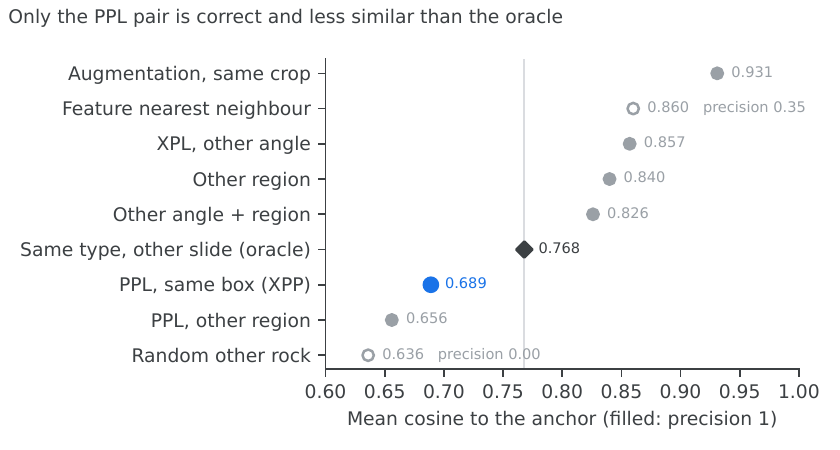}
\caption{Positive sources on thin sections: mean cosine to anchor (ImageNet ResNet-50). Filled markers have precision 1; vertical line marks oracle (another slide of same rock type). Only PPL view of same box is both correct and less similar than oracle.}
\label{fig:sources}
\end{figure}

Thin sections show the same pattern as natural images. The label-free nearest neighbour is wrong 65\% of the time, yet it is more similar to the anchor (0.860) than a true same-type slide (0.768). The acquisition positives are all correct, but most of them are too similar: other rotation angles and other regions lie between augmentation (0.931) and the oracle. Only the PPL view falls below the oracle, and it does so at perfect precision (0.689 at the same location).

Low cosine is a necessary condition, not a sufficient one. The PPL and XPL views show the same grains under different light, not another specimen of the rock, and PPL from another region (0.656) is nearly as far from the anchor as a random other rock (0.636). Whether this variation transfers to cross-slide recognition, or only teaches invariance to illumination, is decided by training: the same-location pair helps (Section~\ref{sec:6.3}), while adding other regions hurts. On thin sections, neighbour pseudo-tasks also fail to improve on instance training on the primary split (Section~\ref{sec:6.2}), as the diagnosis predicts, and remain below XPP on every split (Section~\ref{sec:6.5}).

\section{Method: cross-polarization pairing (XPP)}\label{sec:5}
XPP builds each pseudo-class from one training slide and draws half of its shots from the PPL image at the same field location as an XPL anchor, so the only way to match shots is through what survives the change of illumination. It uses no clustering and no neighbour selection, so the false-positive mechanism described in Section~\ref{sec:4} never arises.

\begin{figure*}[t]
\centering
\includegraphics[width=1.00\textwidth]{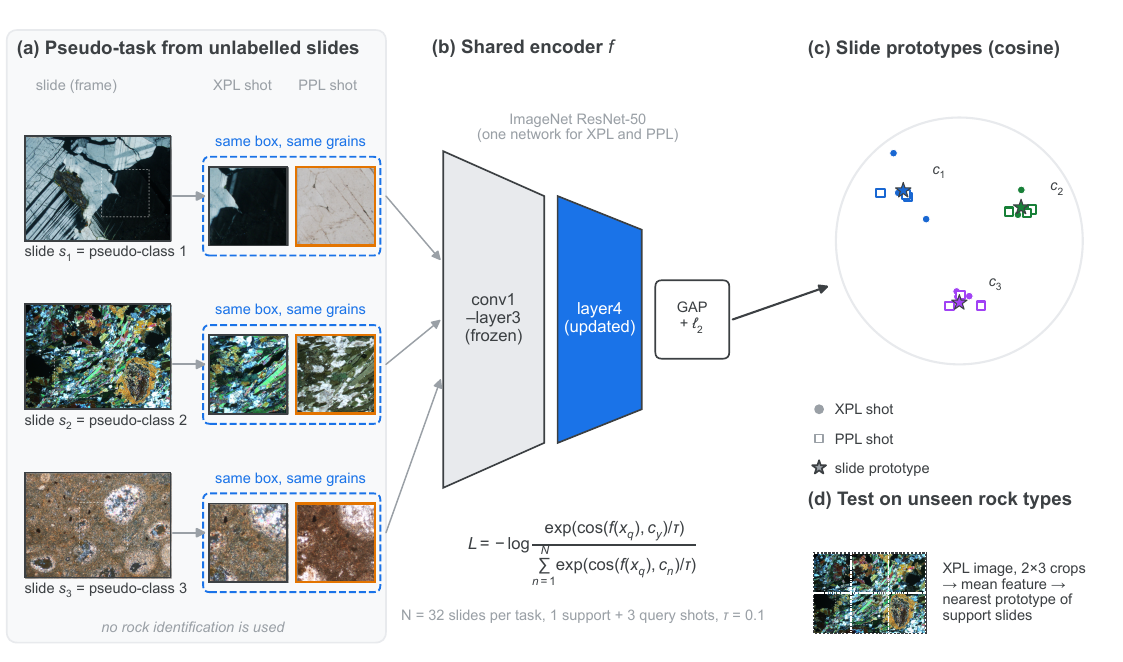}
\caption{XPP framework, drawn with real slides from Nanjing collection (igneous, metamorphic, sedimentary). (a) Each unlabelled slide is one pseudo-class; its shots are one box cropped from aligned XPL frame and from PPL frame of same field. (b) One ImageNet-pretrained ResNet-50 embeds both views; conv1--layer3 are frozen and only layer4 is updated. (c) Shots are pulled to normalized mean of their slide's support shots by a cosine softmax over 32 slides of each task. (d) At test time, an XPL image is described by mean feature of a 2 $\times$ 3 grid of crops and assigned to nearest prototype built from another slide. Network is standard; contribution lies in how pseudo-tasks are built.}
\label{fig:framework}
\end{figure*}

\subsection{Pseudo-classes are slides}\label{sec:5.1}
A pseudo-task samples $N = 32$ training slides as its $N$ ways, with $K = 1$ support and $Q = 3$ query shots per way. A false negative occurs when two sampled slides show the same rock type. Section~\ref{sec:4.1} shows that false negatives cost almost nothing, so we accept them rather than cluster slides, which would reintroduce false positives.

\subsection{Paired views}\label{sec:5.2}
For slide $s$, an anchor XPL image (one of the seven rotation angles) and an anchor box (a square of 0.3--0.6 of the image's short side) are drawn once per task. Every shot of $s$ is then:

\begin{itemize}
  \item with probability 1/2, the anchor box in the \textbf{same XPL image}, or
  \item with probability 1/2, the anchor box in the \textbf{PPL image of the same field}.
\end{itemize}

Each shot is a random sub-crop of 60--100\% of the box area, resized to 224 px, with random 90$^\circ$ rotations and flips and mild brightness and contrast jitter ($\pm$20\%). Hue and saturation are not changed, because colour carries diagnostic information (interference colours, pleochroism). The PPL and XPL frames of a field are geometrically aligned, so the same box covers the same grains.

The box is kept fixed on purpose. Positives from other regions of the slide, or from other rotation angles, reduce accuracy (Section~\ref{sec:6.3}). Other regions share only slide-level properties such as overall tint and magnification, which teaches the model to recognize the slide rather than the rock.

\subsection{Objective}\label{sec:5.3}
Shots are embedded and normalized, prototypes are the normalized mean of the support shots, and queries are classified by cosine similarity with temperature $\tau = 0.1$:

\begin{equation}
\mathcal{L} = -\log \frac{\exp\left(\cos(f(x_q), c_y)/\tau\right)}{\sum_{n=1}^{N} \exp\left(\cos(f(x_q), c_n)/\tau\right)}, \qquad c_n = \frac{\sum_k f(x_{n,k})}{\left\| \sum_k f(x_{n,k}) \right\|}
\end{equation}

\subsection{Encoder and training}\label{sec:5.4}
The encoder is an ImageNet-pretrained ResNet-50, the same backbone family as \citet{wang2025rock}, with the 2048-d pooled feature. Only the last stage (layer4) is fine-tuned, with AdamW (learning rate 1e-5, weight decay 1e-4), a cosine schedule and 1500 iterations. Every 100 iterations, the checkpoint is scored on 500 cross-slide 1-shot episodes of the 10 validation rock types, and the best is kept. Training takes about 6 minutes on one RTX 4090.

Gentle adaptation is essential. Fine-tuning the last two stages at learning rate 1e-4 for 4000 iterations drives every label-free variant below the ImageNet features (Section~\ref{sec:6.4}).

\subsection{Inference}\label{sec:5.5}
At test time a shot is one XPL image. Its feature is the mean of the normalized features of a 2 $\times$ 3 grid of square crops (side 0.5 of the short side, resized to 224 px), matching the training scale. Queries are assigned to the prototype with the highest cosine similarity.

\section{Experiments}\label{sec:6}
On the primary split, XPP is the best label-free method on every protocol and matches a supervised model trained on 78 identified rock types. All numbers are mean $\pm$ standard deviation over three seeds unless stated.

\subsection{Baselines}\label{sec:6.1}
All trained methods share the encoder, the fine-tuning budget of Section~\ref{sec:5.4}, the validation-based checkpoint selection and the test episodes:

\begin{itemize}
  \item \textbf{ImageNet features}: no training on thin sections.
  \item \textbf{Instance training}: pseudo-class = slide, every shot an augmentation of the same XPL box (XPP without the PPL shots).
  \item \textbf{Colour-jitter control}: instance training plus SimCLR-style colour jitter (hue shift up to $\pm$180$^\circ$, saturation $\times$0.3--1.7, 20\% grayscale). It tests whether the PPL view acts as a colour augmentation.
  \item \textbf{Neighbour pseudo-tasks}: the PETD idea on thin sections. With probability rising to 0.6, a shot is taken from the slide's nearest other slide in the current feature space.
  \item \textbf{Supervised}: pseudo-class = identified rock type, using all 78 training types.
\end{itemize}

\subsection{Main results (primary split)}\label{sec:6.2}
\begin{table}[t]
\centering\small
\caption{Main results on primary split: 5-way accuracy (\%), mean $\pm$ sd over three seeds, 2000 test episodes.}
\label{tab:main}
\resizebox{\linewidth}{!}{%
\begin{tabular}{lccccc}
\toprule
Method & Rock identifications & Cross-slide 1-shot & Cross-slide 5-shot & Image-level 1-shot & Image-level 5-shot \\ \midrule
ImageNet features & 0 & 84.96 & 86.43 & 89.49 & 97.31 \\
Instance training & 0 & 87.32 $\pm$ 0.19 & 88.27 $\pm$ 0.20 & 91.19 $\pm$ 0.14 & 97.65 $\pm$ 0.05 \\
Colour-jitter control & 0 & 84.50 $\pm$ 0.18 & 85.83 $\pm$ 0.26 & 88.99 $\pm$ 0.24 & 95.80 $\pm$ 0.09 \\
Neighbour pseudo-tasks & 0 & 87.12 $\pm$ 0.47 & 88.06 $\pm$ 0.48 & 91.05 $\pm$ 0.31 & 97.62 $\pm$ 0.05 \\
\textbf{XPP (ours)} & \textbf{0} & \textbf{89.32 $\pm$ 0.27} & \textbf{90.48 $\pm$ 0.30} & \textbf{92.62 $\pm$ 0.19} & \textbf{98.04 $\pm$ 0.09} \\
Supervised & 78 types & 88.85 $\pm$ 0.36 & 89.79 $\pm$ 0.34 & 92.24 $\pm$ 0.25 & 96.83 $\pm$ 0.10 \\
\citet{wang2025rock}, reported & 78 types & -- & -- & -- & 96.70 \\
\bottomrule
\end{tabular}}
\end{table}
\begin{figure}[t]
\centering
\includegraphics[width=0.80\linewidth]{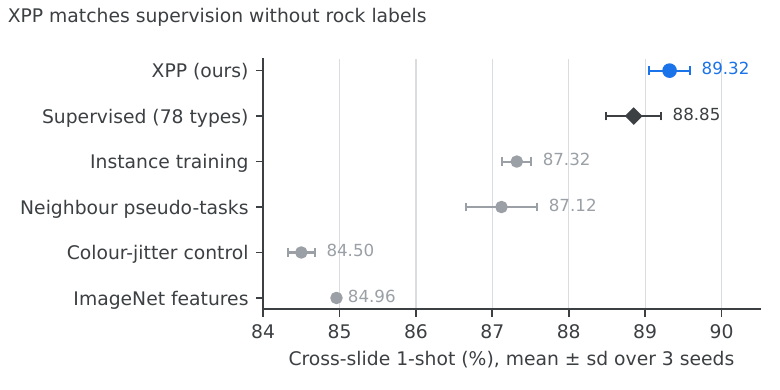}
\caption{Cross-slide 5-way 1-shot accuracy on primary split (mean $\pm$ sd over three seeds). Diamonds use rock identifications.}
\label{fig:mainfig}
\end{figure}

XPP improves on instance training by 2.0 points at cross-slide 1-shot, and on every seed (+1.79, +1.97, +2.27). It improves on ImageNet features by 4.4 points. Without any rock identification it is on par with the supervised model (+0.47 at 1-shot, +0.69 at 5-shot), whose validation accuracy is higher (95.2 against 92.6), so we do not claim it is better.

On the primary split, the colour-jitter control is 4.8 points below XPP and 2.8 below instance training: changing colour is not what helps; pairing two physical views of the same grains is. Across splits, the effect of colour jitter on instance training changes sign, but it stays 1.8 points below XPP on average (Section~\ref{sec:6.5}). On the primary split, neighbour pseudo-tasks do not beat instance training, although slide-level neighbour precision is 0.58, consistent with Section~\ref{sec:4}.

The image-level protocol is saturated: untrained ImageNet features already reach 97.31\% at 5-shot. The reported 96.70\% of \citet{wang2025rock} (different split) is in the same range, so image-level numbers mostly measure re-identification of a field of view.

\subsection{Ablation: which acquisition positives help (primary split)}\label{sec:6.3}
\begin{table}[t]
\centering\small
\caption{Acquisition positives added to instance training (primary split, cross-slide 5-way 1-shot, \%).}
\label{tab:ablation}
\begin{tabular}{lcc}
\toprule
Shots drawn from & Cross-slide 1-shot & vs instance training \\ \midrule
Same XPL box (instance training) & 87.32 $\pm$ 0.19 & -- \\
+ other regions of the slide & 86.11 $\pm$ 0.31 & $-$1.21 \\
+ other rotation angles, same box & 86.89 $\pm$ 0.24 & $-$0.42 \\
+ other regions + other angles & 85.69 $\pm$ 0.08 & $-$1.63 \\
\textbf{+ PPL, same box (XPP)} & \textbf{89.32 $\pm$ 0.27} & \textbf{+2.00} \\
+ PPL + other regions + other angles & 87.85 $\pm$ 0.38 & +0.53 \\
\bottomrule
\end{tabular}
\end{table}
\begin{figure}[t]
\centering
\includegraphics[width=0.80\linewidth]{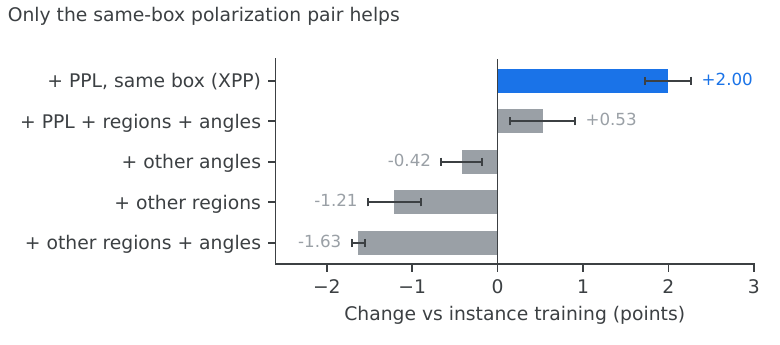}
\caption{Change in cross-slide 1-shot accuracy when acquisition positives are added to instance training (primary split, three seeds).}
\label{fig:ablation}
\end{figure}

Only the polarization pair helps, and adding regions and angles to it costs 1.47 points (every seed lower). Other regions share slide-level appearance but not grains, so they reward recognizing the slide. Other angles change interference colours but keep everything else, which seems too little to force new invariances. The PPL view changes colour completely while keeping every grain boundary, the combination that Section~\ref{sec:4.2} measured as correct (precision 1) and dissimilar (cosine 0.69).

\subsection{Gentle versus aggressive adaptation (primary split)}\label{sec:6.4}
\begin{table}[t]
\centering\small
\caption{Gentle versus aggressive label-free adaptation (primary split, cross-slide 5-way 1-shot, \%). Gentle: mean of three seeds; aggressive: one seed.}
\label{tab:gentle}
\begin{tabular}{lcc}
\toprule
 & Gentle & Aggressive \\
Method & (layer4, lr 1e-5, 1500 it) & (layers 3--4, lr 1e-4, 4000 it) \\ \midrule
Instance training & 87.32 & 76.03 \\
Other regions + angles & 85.69 & 80.42 \\
PPL + regions + angles & 87.85 & 79.50 \\
Neighbour pseudo-tasks & 87.12 & 82.36 \\
Supervised & 88.85 & 89.46 \\
\bottomrule
\end{tabular}
\end{table}
\begin{figure}[t]
\centering
\includegraphics[width=0.80\linewidth]{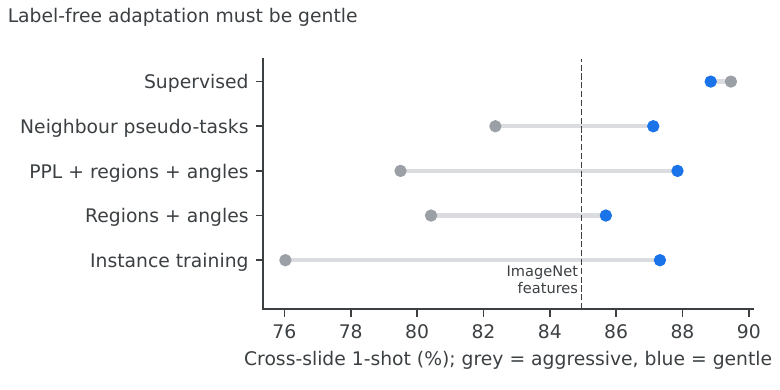}
\caption{Gentle (blue) versus aggressive (grey) adaptation. Dashed line marks ImageNet features without training.}
\label{fig:gentle}
\end{figure}

Aggressive fine-tuning reaches near-100\% pseudo-task accuracy within 1000 iterations and pushes all label-free variants below the ImageNet features (84.96), while the supervised model is unaffected. Slide-level pseudo-tasks are easy to solve with superficial cues, so label-free adaptation must stay close to the pretrained features.

\subsection{Robustness across random splits}\label{sec:6.5}
\begin{table}[t]
\centering\small
\caption{Cross-slide 5-way 1-shot accuracy (\%) on three random rock-type splits.}
\label{tab:splits}
\resizebox{\linewidth}{!}{%
\begin{tabular}{lcccccc}
\toprule
Split & ImageNet & Instance & Colour jitter & Neighbour & XPP & Supervised \\ \midrule
0 (primary) & 84.96 & 87.32 $\pm$ 0.19 & 84.50 $\pm$ 0.18 & 87.12 $\pm$ 0.47 & \textbf{89.32 $\pm$ 0.27} & 88.85 $\pm$ 0.36 \\
1 & 85.35 & 85.39 $\pm$ 0.36 & 88.89 $\pm$ 0.25 & 89.56 $\pm$ 0.39 & \textbf{89.87 $\pm$ 0.10} & 88.57 $\pm$ 0.38 \\
2 & 81.58 & 84.36 $\pm$ 0.28 & 85.34 $\pm$ 0.14 & 84.29 $\pm$ 0.16 & 85.04 $\pm$ 0.62 & 87.19 $\pm$ 0.20 \\ \midrule
Mean of splits & 83.96 & 85.69 & 86.24 & 86.99 & \textbf{88.08} & 88.20 \\
\bottomrule
\end{tabular}}
\end{table}

XPP improves on ImageNet features on every split (+4.36, +4.52 and +3.46 points) and on instance training on every split (+2.00, +4.48 and +0.68). Averaged over the three splits, it reaches 88.08\%, the highest of the label-free methods (instance training 85.69, colour jitter 86.24, neighbour pseudo-tasks 86.99) and on par with the supervised model (88.20). The ranking of the other methods is less stable. Instance training barely improves on ImageNet features on split 1 (+0.04), and colour jitter, which hurts instance training on the primary split, helps it on splits 1 and 2. On split 2, colour jitter is 0.30 points above XPP, and the supervised model leads by 2.15 points.

With 20 test rock types, the choice of split moves even the untrained baseline by 3.8 points. Single-split results in this domain, including prior work, should therefore be read with that variation in mind. We report the primary split as the main benchmark, as is standard, and give these numbers so that the size of the effect can be judged.

\subsection{Label efficiency (primary split)}\label{sec:6.6}
\begin{table}[t]
\centering\small
\caption{Label efficiency on primary split: supervised meta-training with $k$ identified rock types versus XPP (\%).}
\label{tab:labeleff}
\begin{tabular}{lccc}
\toprule
Training signal & Rock identifications & Cross-slide 1-shot & Cross-slide 5-shot \\ \midrule
ImageNet features & 0 & 84.96 & 86.43 \\
Supervised & 10 types (30 slides) & 85.78 $\pm$ 0.09 & 87.15 $\pm$ 0.08 \\
Supervised & 20 types & 86.13 $\pm$ 0.31 & 87.04 $\pm$ 0.32 \\
Supervised & 40 types & 87.22 $\pm$ 0.23 & 88.27 $\pm$ 0.30 \\
Supervised & 78 types (all) & 88.85 $\pm$ 0.36 & 89.79 $\pm$ 0.34 \\
\textbf{XPP} & \textbf{0} & \textbf{89.32 $\pm$ 0.27} & \textbf{90.48 $\pm$ 0.30} \\
\bottomrule
\end{tabular}
\end{table}
\begin{figure}[t]
\centering
\includegraphics[width=0.80\linewidth]{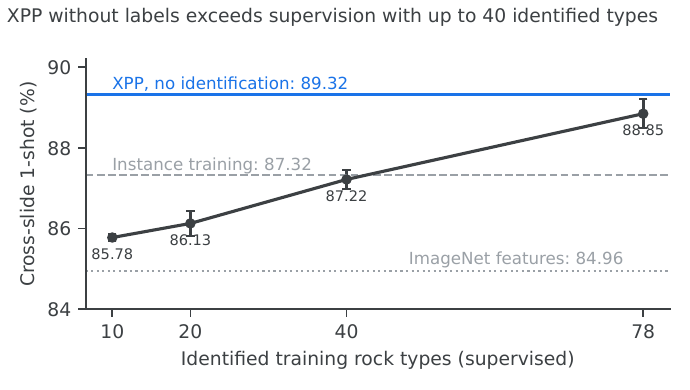}
\caption{Supervised meta-training with $k$ identified rock types (black) compared with XPP, which uses none (blue line).}
\label{fig:labeleff}
\end{figure}

Without any identification, XPP is above every supervised model trained on a subset of the rock types and on par with the one trained on all 78. Instance training (87.32) matches the supervised model with 40 identified types. The 10 identified types are drawn at random from the training types.

\subsection{What XPP learns (primary split, seed 1)}\label{sec:6.7}
\begin{table}[t]
\centering\small
\caption{Cross-illumination alignment of encoders (primary split, seed 1, 20 unseen test rock types). Retrieval: share of PPL images whose nearest XPL image among all test fields is their own field.}
\label{tab:invariance}
\begin{tabular}{lccc}
\toprule
 & \multicolumn{2}{c}{PPL--XPL cosine} & PPL retrieves its \\ \cmidrule(lr){2-3}
Encoder & Same field & Other fields & XPL field (top-1) \\ \midrule
ImageNet & 0.743 & 0.636 & 25.0\% \\
Instance training & 0.740 & 0.615 & 16.7\% \\
Colour-jitter control & 0.699 & 0.460 & 33.3\% \\
\textbf{XPP} & \textbf{0.819} & 0.651 & \textbf{50.0\%} \\
Supervised & 0.674 & 0.491 & 13.3\% \\
\bottomrule
\end{tabular}
\end{table}

\begin{table}[t]
\centering\small
\caption{Recognition accuracy of same encoders (primary split, seed 1, \%). Greyscale: test images converted to greyscale. 20-way: all test rock types at once, prototypes from other slides.}
\label{tab:recognition}
\begin{tabular}{lccc}
\toprule
 & Cross-slide & Cross-slide 1-shot, & 20-way, leave- \\
Encoder & 1-shot & greyscale images & one-slide-out \\ \midrule
ImageNet & 84.96 & 83.55 & 75.95 \\
Instance training & 87.23 & 88.34 & 79.29 \\
Colour-jitter control & 84.34 & 85.83 & -- \\
\textbf{XPP} & \textbf{89.02} & \textbf{88.72} & \textbf{85.71} \\
Supervised & 88.80 & 85.43 & -- \\
\bottomrule
\end{tabular}
\end{table}
\begin{figure}[t]
\centering
\includegraphics[width=1.00\linewidth]{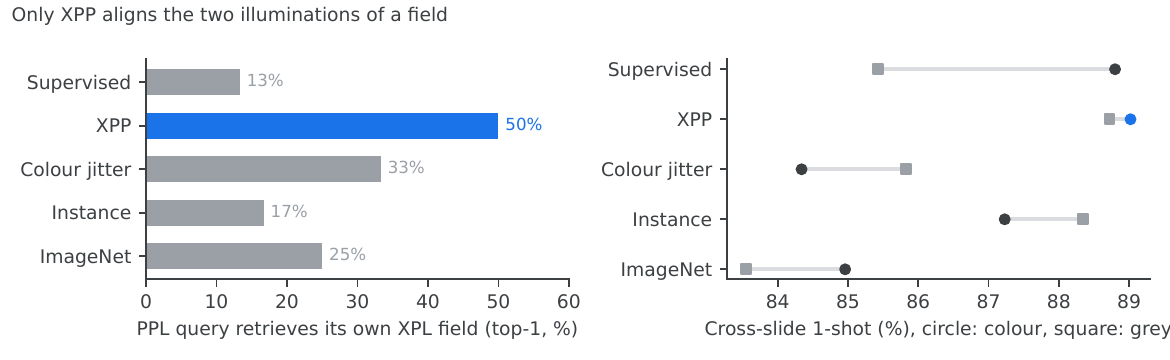}
\caption{Left: share of PPL images whose nearest XPL image among all test fields is their own field. Right: cross-slide 1-shot accuracy on original (coloured dot) and greyscale (grey dot) test images.}
\label{fig:invariance}
\end{figure}

XPP is the only encoder that brings the two illuminations of a field together (Table~\ref{tab:invariance}): same-field PPL--XPL cosine rises from 0.74 to 0.82, and a PPL image retrieves its own XPL field among all test fields twice as often as with ImageNet features. In a harder 20-way task (Table~\ref{tab:recognition}) (every test rock type at once, prototypes from the other slides), XPP reaches 85.71\% against 79.29\% for instance training.

Removing colour at test time costs the supervised model 3.4 points, but neither label-free model (XPP $-$0.3, instance training +1.1). Robustness to colour is therefore a property of slide-level self-supervision in general, not of the polarization pair alone; what the pair adds is cross-illumination alignment. Grad-CAM maps (layer4, 14 $\times$ 14) of the three encoders are nearly identical, as expected when only the last stage is fine-tuned gently, and are given in the supplementary material.

\section{Discussion}\label{sec:7}
\textbf{Why polarization pairs work when other positives do not.} Section~\ref{sec:4} showed that useful positives must be correct and must differ from the anchor in ways the current features do not already discount. Feature neighbours fail the first condition; augmentations, other angles and other regions fail the second, or differ in the wrong way (slide-level tint and magnification). A PPL--XPL pair changes nearly all colour information while preserving every grain boundary, so the only shared signal is the petrographic structure that also distinguishes rock types across slides.

\textbf{Why acquisition metadata is not a label.} The pairing uses only the fact that two photographs show the same field. It never states what the rock is, which is the expensive, expert step. Any laboratory that captures PPL and XPL views of a field, which is standard practice, can apply XPP to an archive that has never been identified.

\textbf{Why this setting, with about 230 training slides.} Identifying the training slides of a teaching collection would be cheap; the use case is a large digitized archive, a new drilling campaign or a multi-laboratory collection, where identification lags acquisition. The Nanjing collection is the public benchmark on which the gap to supervision can be measured. Only checkpoint selection uses identified samples, a validation set of 10 rock types; the training slides themselves are never identified.

\section{Conclusions}\label{sec:8}
This paper addresses few-shot classification of rock thin sections when no training slide has been identified, and the only information available is the acquisition metadata recorded by the microscope. We analysed why generic unsupervised pseudo-tasks fit this setting poorly and proposed cross-polarization pairing (XPP), which uses the PPL and XPL views of the same field as positives. Evaluated on the Nanjing University collection under a cross-slide protocol, the work leads to the following conclusions and future directions.

(1) Generic unsupervised pseudo-tasks are limited by their positives. Oracle experiments show that false positives, not false negatives, cost accuracy, and that label-free selection returns positives that are either wrong or near-duplicates of the anchor. On thin sections, only the PPL view of the same field is both correct and less similar to the anchor than another slide of the same rock type.

(2) Cross-polarization pairing turns this property of polarized-light microscopy into a training signal. Without any rock identification, XPP gains 4.4 points over ImageNet features and 2.0 points over instance training on the primary split, and matches a supervised model meta-trained on 78 identified rock types, on the primary split and on average over three random splits.

(3) The gain comes from the pairing itself. Strong colour augmentation, other regions of the slide and other rotation angles do not reproduce it, and XPP is the only encoder that aligns the two illuminations of a field while staying robust to the removal of colour.

(4) The evaluation protocol strongly affects reported accuracy. Under the image-level protocol of earlier work, untrained ImageNet features already reach 97\% 5-shot accuracy, whereas the cross-slide protocol measures recognition of a rock type rather than of a field of view. With 20 test rock types, the choice of split alone moves accuracy by several points, so results in this domain should be reported over several splits.

(5) The size of the gain over instance training varies between random splits, and the collection offers one field of view per slide. In future work, we plan to apply XPP to larger, multi-laboratory archives with several fields per slide and full rotation series, to extend it to mineral-scale recognition, and to remove the small identified validation set now used for checkpoint selection.

\section*{Code availability}
The source code, experiment scripts and documentation are available at \url{https://github.com/waterfall132/CPP} (MIT license).

\section*{Data availability}
The thin-section photographs are the public Nanjing University collection \citep{lai2020photomicrograph}. The rock-type splits and fixed test episodes used in this work are included in the code repository.

\section*{Declaration of competing interest}
The authors declare that they have no known competing financial interests or personal relationships that could have appeared to influence the work reported in this paper.

\section*{Declaration of generative AI and AI-assisted technologies in the manuscript preparation process}
During the preparation of this work the authors used ChatGPT (OpenAI) to assist with editing the text and preparing the LaTeX manuscript. After using this tool, the authors reviewed and edited the content as needed and take full responsibility for the content of the article.

\section*{Acknowledgments}
This work was supported by the Innovation Project of GUET Graduate Education (2024YCXS048).

\appendix
\section{Oracle diagnosis on miniImageNet}\label{app:diag}
\setcounter{table}{0}\setcounter{figure}{0}\setcounter{equation}{0}
\renewcommand{\thetable}{A.\arabic{table}}\renewcommand{\thefigure}{A.\arabic{figure}}\renewcommand{\theequation}{A.\arabic{equation}}

\paragraph{Setup.} The base method is PETD \citep{zhao2025unsupervised}: online k-means with 512 clusters, and progressive replacement of query views by co-clustered samples with probability $0.6\cdot t/T$. It uses a Conv4 encoder, 5-way 1-shot 5-query pseudo-tasks, raw-resolution images and $T=100$ epochs unless stated. Models are selected on the validation classes, and all configurations are evaluated on the same 10\,000 fixed 5-way test episodes, so that they can be compared episode by episode; the 95\% intervals over episodes are about $\pm$0.2. Results are from a single seed. The baseline, no-replacement and oracle runs were trained on a different machine from the selector and precision runs; differences between machines are a few tenths of a point, which is larger than the episode-level interval.

\paragraph{False positives and false negatives.} Table~\ref{tab:fn} separates the two errors with oracles. About 90\% of PETD's injected neighbours belong to another class. Removing them adds 3.1 points at 1-shot and 4.5 at 5-shot; at 5-shot, PETD is even 1.4 points below using no neighbours at all. Removing false negatives changes accuracy by less than 0.4 points (+0.34 and +0.31). The false-positive gap is not an artefact of short training: with $T=400$, PETD reproduces the published accuracy (50.74 / 66.34 against 50.46 / 66.40), and removing false positives still adds 3.5 and 4.3 points (54.20 / 70.59).

\begin{table}[h]
\centering\small
\caption{Oracle decomposition of pseudo-task errors of PETD on miniImageNet (Conv4, $T=100$, 5-way test accuracy, \%).}
\label{tab:fn}
\begin{tabular}{lccc}
\toprule
 & Injected positives & & \\
Setting & correct & 1-shot & 5-shot \\ \midrule
No replacement (own views only) & -- & 48.71 & 65.87 \\
PETD & about 10\% & 49.47 & 64.47 \\
PETD, false positives removed by oracle & 100\% & 52.61 & 68.94 \\
PETD, false negatives removed by oracle & about 10\% & 49.81 & 64.78 \\
\bottomrule
\end{tabular}
\end{table}

False negatives are rare in PETD (about 15\% of tasks contain two ways of the same class), so we also test ProtoCLR \citep{medina2020self}, an augmentation-based method in which every image is its own class and false negatives grow with the number of training ways (Table~\ref{tab:protoclr}). Even at 50 ways, where a batch contains about 19 same-class pairs, removing them changes 1-shot accuracy by $-0.32$ points, and the differences across $N$ have no consistent sign. At $N=50$, only about 1.6\% of the negatives of an image share its class. The number of ways itself matters: 1-shot accuracy rises by 2.8 points from $N=5$ to $N=50$. Our evaluation of the released ProtoCLR checkpoint gives 44.12 / 62.43 against the published 44.89 / 63.35.

\begin{table}[h]
\centering\small
\caption{False negatives in ProtoCLR on miniImageNet (Conv4, 84$\times$84 images, 150\,000 iterations, single seed, 5-way 1-shot test accuracy, \%).}
\label{tab:protoclr}
\begin{tabular}{ccccc}
\toprule
Training & Same-class pairs & & False negatives & \\
ways $N$ & per batch & ProtoCLR & removed & Difference \\ \midrule
5  & 0.16  & 40.89 & 41.17 & $+0.28$ \\
10 & 0.70  & 42.37 & 42.29 & $-0.07$ \\
20 & 2.96  & 42.98 & 43.20 & $+0.22$ \\
50 & 19.12 & 43.68 & 43.36 & $-0.32$ \\
\bottomrule
\end{tabular}
\end{table}

\begin{figure}[h]
\centering
\includegraphics[width=0.80\linewidth]{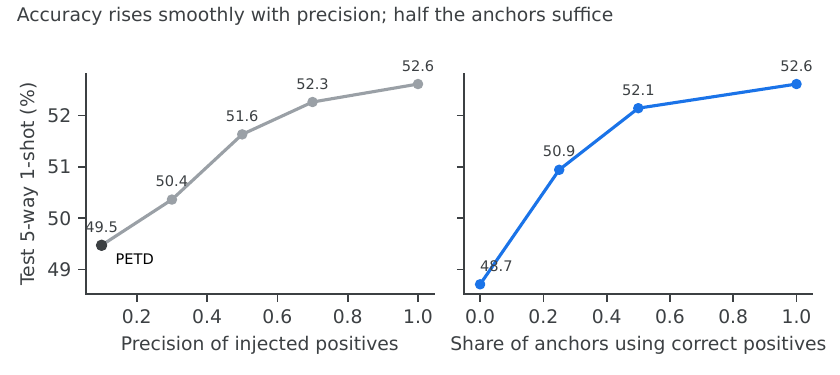}
\caption{Test 5-way 1-shot accuracy of PETD on miniImageNet when injected positives are replaced by an oracle. Left: precision of injected positives at full coverage; right: share of anchors that receive correct positives.}
\label{fig:dose}
\end{figure}

\paragraph{Precision and coverage.} Accuracy rises smoothly with the precision of injected positives (Table~\ref{tab:dose}, Fig.~\ref{fig:dose}), by about 0.5 points per 0.1 of precision up to 0.5, and saturates above 0.7; there is no threshold. Below a precision of 0.3, injection hurts 5-shot accuracy. Correct positives for half of the anchors recover 88\% of the 1-shot gain and all of the 5-shot gain.

\begin{table}[h]
\centering\small
\caption{Accuracy as a function of precision of injected positives and of share of anchors that receive them (miniImageNet, $T=100$, \%). Precision is controlled with class labels.}
\label{tab:dose}
\begin{tabular}{lccc}
\toprule
Precision of injected positives & Anchors using positives & 1-shot & 5-shot \\ \midrule
0.30 & 100\% & 50.36 & 65.78 \\
0.50 & 100\% & 51.63 & 67.23 \\
0.70 & 100\% & 52.26 & 68.07 \\
1.00 & 25\% & 50.94 & 67.78 \\
1.00 & 50\% & 52.14 & 68.84 \\
\bottomrule
\end{tabular}
\end{table}

\paragraph{Label-free selectors.} Table~\ref{tab:select} replaces PETD's random co-clustered neighbour with label-free selectors. An episodic prototype gate does separate neighbours (precision 0.22 among kept, 0.07 among rejected), but no selector raises precision above 0.4, and all land between 49.2 and 49.8 at 1-shot. Restricting PETD's own neighbours to the densest half of the anchors does not help either (49.71), so reducing coverage alone is not the issue. At matched precision (0.31 against 0.30), global nearest neighbours trail the precision oracle of Table~\ref{tab:dose} by 0.6 points.

\begin{table}[h]
\centering\small
\caption{Label-free neighbour selectors on miniImageNet ($T=100$). Precision is measured on neighbours actually injected; replaced share is for final epoch.}
\label{tab:select}
\resizebox{\linewidth}{!}{%
\begin{tabular}{lcccc}
\toprule
Label-free selector & Precision of used neighbours & Query slots replaced (final epoch) & 1-shot & 5-shot \\ \midrule
PETD (random co-clustered) & about 0.10 & 60\% & 49.47 & 64.47 \\
Nearest co-clustered (kNN) & 0.18 & 60\% & 49.76 & 65.83 \\
Episodic prototype gate & 0.22 & 17\% & 49.50 & 66.20 \\
kNN + gate & 0.25 & 32\% & 49.56 & 66.30 \\
Global nearest neighbours & 0.31 & 60\% & 49.74 & 66.59 \\
Global NN, densest 50\% of anchors & 0.39 & 30\% & 49.17 & 66.09 \\
PETD, densest 50\% of anchors & 0.16 & 30\% & 49.71 & 65.84 \\
\bottomrule
\end{tabular}}
\end{table}

\paragraph{Why selection fails.} An offline audit of a trained checkpoint scores every co-clustered candidate of each anchor. The best signals (cosine, reciprocal kNN) reach AUROC 0.75 and top-5 precision 0.22, against a base rate of 0.10; averaging features over 8 augmentations does not help (0.22). Without the cluster restriction, the global top-5 neighbours reach precision 0.38. Graph diffusion on the kNN graph (personalized PageRank, $k\in\{10,20\}$, $\alpha\in\{0.8,0.95,0.99\}$) gives precision 0.36--0.38 at mean cosine 0.70--0.71. Skipping the nearest 5, 20 or 50 candidates lowers cosine only to 0.65--0.69, while precision drops to 0.24--0.32. Random true same-class samples sit at cosine 0.55, whereas the correct neighbours found by label-free selectors sit at 0.72. Synthesizing harder positives from a single image, for example by feature extrapolation, does not help either: it adds no information about other instances of the class, and own-view training is the lowest 1-shot entry in Table~\ref{tab:fn}.

\bibliographystyle{cas-model2-names}
\bibliography{refs}

\end{document}